\documentclass[letterpaper, 10 pt, journal]{IEEEtran}
\usepackage{amsmath,amsfonts}
\usepackage{textcomp}
\usepackage{stfloats}
\usepackage{verbatim}
\usepackage{graphicx}
\usepackage{url}
\usepackage{pifont}
\usepackage{cases}
\usepackage{multirow}
\usepackage{ulem}
\usepackage{bbm}
\usepackage{arydshln}
\def\BibTeX{{\rm B\kern-.05em{\sc i\kern-.025em b}\kern-.08em
    T\kern-.1667em\lower.7ex\hbox{E}\kern-.125emX}}
\usepackage{balance}
\usepackage{cite}
\usepackage{booktabs}
\begin{document}
\title{PAGCNet: A Pose-Aware and Geometry Constrained Framework for Panoramic Depth Estimation}
\author{Kanglin Ning, Ruzhao Chen, Penghong Wang, Xingtao Wang, Ruiqin Xiong, \textit{Senior Member, IEEE}, Xiaopeng Fan, \textit{Senior Member, IEEE}
\thanks{This work was supported in part by the National Key R\&D Program of China (2023YFA1008500), the National Natural Science Foundation of China (NSFC) under grants 62402138 and U22B2035. (Corresponding author: Xiaopeng Fan.)}
\thanks{Kanglin Ning, Ruzhao Cheng, Penghong Wang, Xingtao Wang are with the Faculty of Computing, Harbin Institute of Technology, Harbin 150001, China. Kanglin Ning, Penghong Wang and Xingtao Wang are also currently affiliated with the Suzhou Research Institute of HIT. (email: 23B936010@stu.hit.edu.cn; 24S103291@stu.hit.edu.cn;phwang@hit.edu.cn; xtwang@hit.edu.cn)}
\thanks{R. Xiong is with the Institute of Digital Media, School of Electronic Engineering and Computer Science, Peking University, Beijing 100871, China (e-mail: rqxiong@pku.edu.cn)}
\thanks{Xiaopeng Fan is with the Faculty of Computing, Harbin Institute of Technology, Harbin 150001, China. He is also with the PengChengLab, Shenzhen 518055, China, and the Suzhou Research Institute of HIT. (e-mail: fxp@hit.edu.cn).}
}

\markboth{IEEE Transactions on Multimedia}%
{PAGCNet: A Pose-Aware and Geometry Constrained Framework for Panoramic Depth Estimation}

\maketitle

\begin{abstract}
Explicitly modeling room background depth as a geometric constraint has proven effective for panoramic depth estimation. However, reconstructing this background depth for regular enclosed regions in a complex indoor scene without external measurements remains an open challenge. To address this, we propose a pose-aware and geometry-constrained framework for panoramic depth estimation. Our framework first employs multiple task-specific decoders to jointly estimate room layout, camera pose, depth, and region segmentation from a input panoramic image. A pose-aware background depth resolving (PA-BDR) component uses tasks decoder's prediction to resolve the camera pose. Subsequently, the proposed PA-BDR component uses the camera pose to compute the background depth of regular enclosed regions and uses this background depth as a strong geometric prior. Based on the output of the region segmentation decoder, a fusion mask generation (FMG) component produces a fusion weight map to guide where and to what extent the geometry-constrained background depth should correct the depth decoder's prediction. Finally, an adaptive fusion component integrates this refined background depth with the initial depth prediction, guided by the fusion weight. Extensive experiments on Matterport3D, Structured3D, and Replica datasets demonstrate that our method achieves significantly superior performance compared to current open-source methods. Code is available at https://github.com/emiyaning/PAGCNet.
\end{abstract}


\section{Introduction}
To enable 3D understanding of indoor environments from a single omnidirectional image, depth estimation becomes a fundamental requirement\cite{lin2025one}. This potential for inferring entire scene structure from a single panorama has motivated active research on panoramic depth estimation.

\begin{figure}[t]
\centering
\includegraphics[width=3.4in]{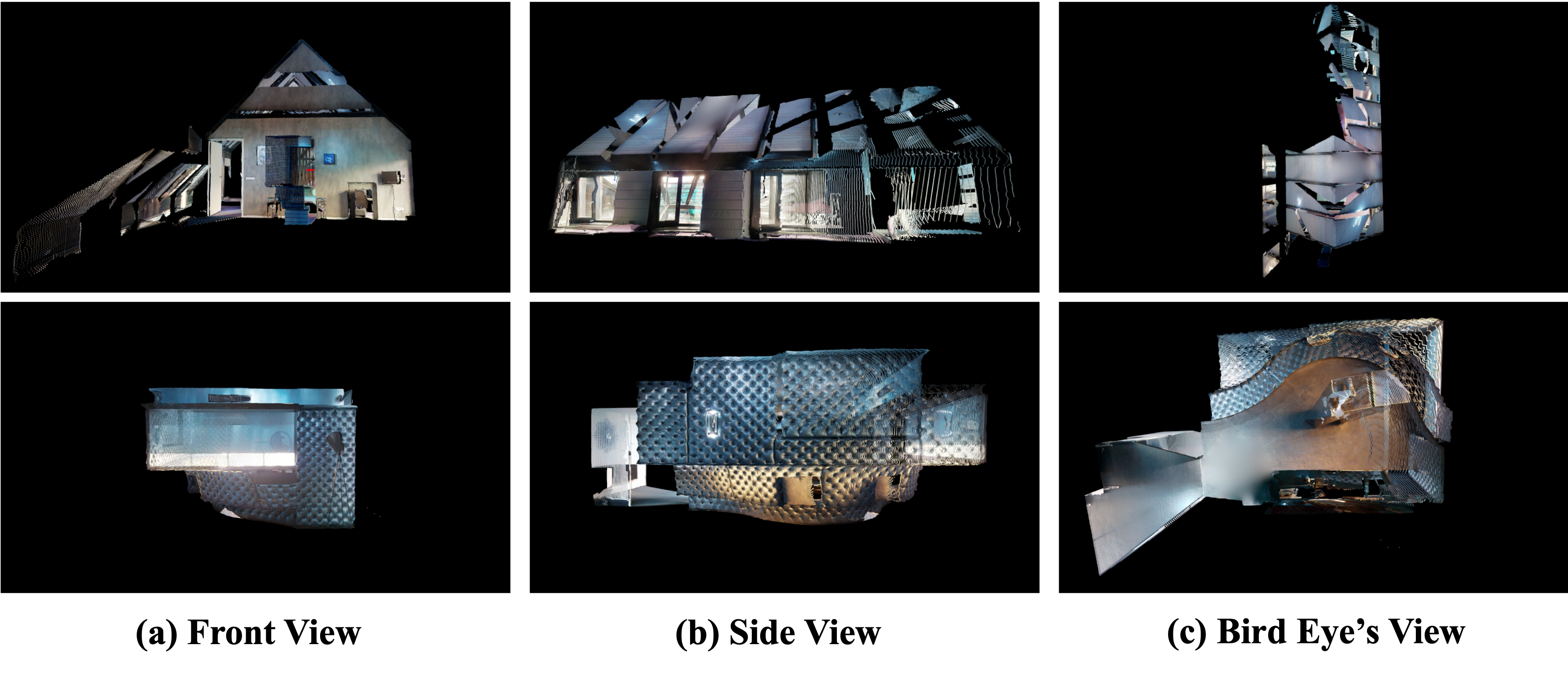}
\caption{This figure presents 3D visualizations of the ground truth depth maps for two rooms with an atypical structure, shown from three distinct viewpoints: frontal, side, and bird’s-eye view (BEV). The first row depicts a room whose overall volume approximates a triangular prism shape. The second row illustrates a KTV room, in which the sofa cushions and the entire wall are merged into a single continuous structure.}
\setlength{\abovecaptionskip}{-0.3cm}
\end{figure}

Existing methods for monocular panoramic depth estimation primarily adopt three technical strategies to address inherent object distortions\cite{gao2022review}. One line of work\cite{shen2022panoformer, yun2023egformer} designs specialized feature extractors, informed by panoramic imaging principles, to directly process distorted equirectangular images. Another common strategy \cite{wang2020bifuse} projects the equirectangular panorama (ERP) onto another projection, extracts features separately from both, and then fuses them to predict the depth. A more recent direction\cite{chen2024bgdnet} explicitly incorporates geometric priors; it uses estimated room layout to construct a background  depth model, which significantly improves accuracy in geometrically discontinuous regions. While these methods achieve strong performance on benchmarks with regular, Manhattan-aligned layouts, they often overlook the prevalence of complex shaped rooms in real-world scenarios.

As illustrated in Fig. 1, real-world rooms are often characterized by irregular, non-Manhattan layouts. In contrast, the background depth estimated by BGDNet \cite{chen2024bgdnet} based on its predicted room layout typically assumes a regular room structure and known camera pose, which limits the method's performance in the scenarios depicted in Fig. 1. Nevertheless, prior work \cite{10657716} has suggested that any irregular room can be decomposed into regular enclosed regions and irregular regions. The room layout annotations provided by various datasets usually correspond to the regular enclosed regions in such scenarios. Hence, the reliable identification and separation of regular and irregular regions in specifically structured indoor scenes, as well as the subsequent reconstruction of background depth for regular enclosed regions without external measurements, remains an open challenge.

To address existing problems in panoramic depth estimation, this paper proposes a depth estimation framework named as pose-aware and room geometry-constrained depth estimation framework (PAGCNet). The proposed framework based on multi-task learning unifying four tasks: depth estimation, camera pose estimation, room layout estimation, and region segmentation. Specifically, region segmentation decoder perform two semantic segmentation task to extract the irregular region mask and background mask from input panorama. The irregular region mask used as an indicator to tell depth estimator which region fall outside the enclosed background region defined by room layout, while background mask used to tell our framework to what extent the geometry-constrained background depth should correct the depth decoder's prediction. Building on this output, our PA-BDR component optimize the camera pose decoder's prediction and further resolving the enclosed region background depth. Then, our FGM component produces the fusion weight map according to the two mask prediction of region segmentation decoder. Finally, an adaptive fusion component integrates the enclosed region background depth with the depth decoder's prediction under the guidance of fusion weight map, yielding a refined depth prediction. Extensive experimental results on the Matterport3D, Structured3D, Replica dataset show that our proposed methods can achieve significantly superior performance than current open-source methods. In general, our contributions can be summarized into the following three points:

\begin{itemize}
\item We proposes PAGCNet, which calculates the enclosed region background depth to adaptively optimizes the final depth prediction.
\item We design a PA-BDR component to compute the background depth without external measurement of camera pose. 
\item We introduce an FMG component and an adaptive fusion component. The FMG component uses the region segmentation decoder's output to compute the fusion weight, while the adaptive fusion component uses this fusion weight to perform the fusion between background depth and depth decoder's prediction.
\end{itemize}

\section{Related Works}
A core challenge in panoramic depth estimation is image distortion. Existing research primarily employs three approaches: 1) estimating depth solely on a single projection; 2) projecting the panorama onto multiple modalities for depth estimation; and 3) using generated background depth to guide estimation.

\subsection{Single Projection Inputs}
Panoramic images employ spherical representations with $180^{\circ}$ vertical and $360^{\circ}$ horizontal fields of view. Processing typically involves projecting these spherical images onto 2D planes through perspective mapping. Common modalities include equirectangular\cite{zioulis2018omnidepth, cheng2020omnidirectional, shen2022panoformer, yun2023egformer, zhang2025sgformer}, tangent\cite{eder2020tangent, li2022omnifusion, rey2022360monodepth}, and icosahedron projections\cite{zhang2019orientation, lee2019spherephd}. 

For depth estimation, equirectangular projection predominates. Omnidepth\cite{zioulis2018omnidepth} introduced RectNet for efficient equirectangular feature extraction and depth prediction. ODE-CNN\cite{cheng2020omnidirectional} proposed a hardware-software co-design system. Following vision transformers' emergence, specialized models have proliferated: Panoformer\cite{shen2022panoformer} developed reference-point window self-attention for equirectangular features within an encoder-decoder architecture; Egformer\cite{yun2023egformer}, SGFormer\cite{zhang2025sgformer}, GLPanoDepth\cite{bai2024glpanodepth} subsequently incorporated global receptive fields and spherical geometry constraints. 

Alternatively, tangent-view approaches \cite{eder2020tangent, li2022omnifusion, rey2022360monodepth} project panoramas onto multiple views, estimate per-view depth, and spatially composite patches into panoramic depth maps. Vertically compressed methods \cite{yu2023panelnet, sun2021hohonet, pintore2021slicenet} adapt layout estimation techniques, reducing panoramas to 1-pixel height and employing Bi-LSTMs or self-attention for depth estimation.

\subsection{Bi-Projection Inputs}
Methods using bi-projection inputs project panorama images onto two distinct perspectives. Of the bi-projection input methods, equirectangular projection is usually used as the primary perspective by default. These methods usually use shared or dedicated branch networks to predict corresponding depth maps. Then, the another perspective-specific depth estimate are reprojected to the equirectangular domain and composited with directly predicted equirectangular depth maps to yield refined depth estimations. Representative examples include BiFuse\cite{wang2020bifuse}, BiFuse++\cite{wang2022bifuse++}, and UniFuse\cite{jiang2021unifuse}, which project onto cube maps; HRDFuse\cite{ai2023hrdfuse} and GA360Fuse\cite{wang2025geometry}, which fuse equirectangular and tangent depth predictions; and Elite360D\cite{ai2024elite360d}, which fuses equirectangular and ICOSAP\cite{lee2019spherephd} perspectives for enhanced accuracy. Intuitively, such methods seem better suited to handle distortion than single-view panoramic depth estimation. However, current public dataset rankings show that supervised learning on pure equirectangular images achieves superior performance. Moreover, dual-view projection inevitably introduces significant additional computational overhead during training and inference. Therefore, the benefits of this approach may not justify its computational cost.

\subsection{Background Based Methods}
Current state-of-the-art methods demonstrate strong quantitative performance across benchmarks. However, their 3D depth visualizations frequently exhibit structural inaccuracies in room geometry and over-smoothed corners. To address this limitation, recent approaches\cite{chen2024bgdnet} integrate room structural priors into panoramic depth estimation. Driven by advances in room layout estimation, current models\cite{sun2019horizonnet, zhao20223d, pintore2020atlantanet, wang2021led2, ai2023hrdfuse, 10657716} demonstrate high accuracy in predicting layouts for rooms with regular, Manhattan-aligned structures. Yet, in real-world indoor images, room structures are often irregular and not based on a Manhattan layout, and the camera's pose itself cannot be obtained in real time. This severely limits its performance in real-world indoor scenes with irregular structures. To eliminate the dependence on these two assumptions, our multi-task learning framework incorporates camera pose learning and irregular region segmentation tasks. Our method is detailed in the following sections.

\begin{figure*}[!t]
\centering
\includegraphics[width=7.0in]{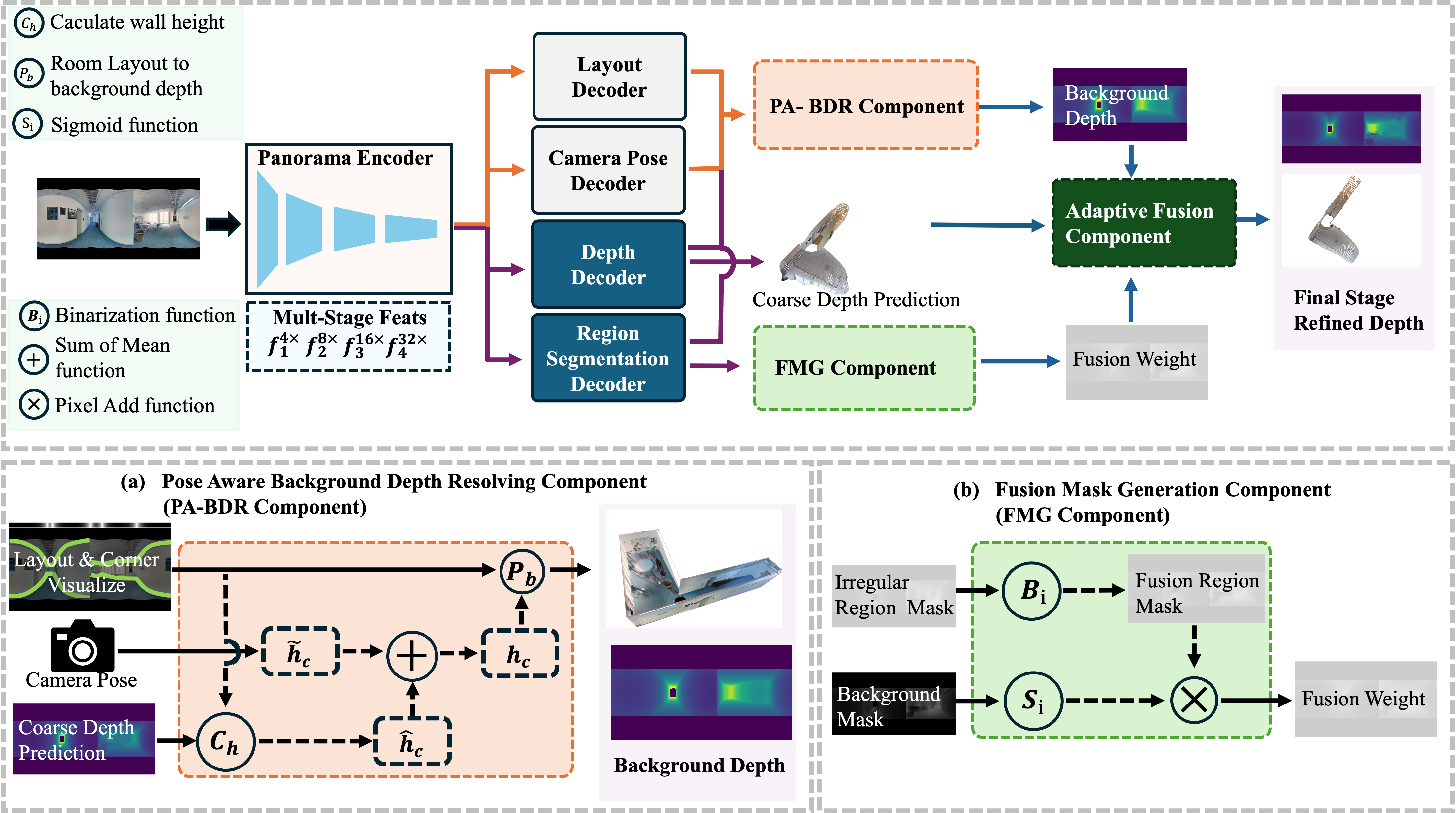}
\caption{The structure diagram of our proposed PAGCNet. In terms of model structure, our framework includes a shared panorama encoder and four task-corresponding decoders. Based on the obtained layout estimation, first stage depth, camera pose, irregular region mask and background region mask, the proposed framework decode fine-grained depth prediction.} 
\end{figure*}

\section{Methodology}
The proposed PAGCNet's diagram in this paper is shown in Fig.2. From the view of network architecture, the framework includes a shared panorama encoder and four task-corresponding decoder modules. Moreover, our framework contain a PA-BDR component, a FGM component and an adaptive fusion component. The panorama encoder extract the multi-stage feature sets $F = \{f^{2^{i+1} \times}_{i} | i=1,2,3,4 \}$﻿ from the panorama image. The proposed PA-BDR component optimize the camera pose and resolving regular enclosed region's background depth based on tasks decoder's prediction. Then, FMG component extract the fusion weight map from input panorama with two sub binary semantic segmentation task's prediction. Finally, the adaptive fusion component will fuse the regular enclosed region's background depth and coarse depth prediction of depth decoder to get the refined depth predictions. In the following subsections we will introduce each module in detail.

\subsection{Architecture}

\textbf{Panorama Encoder: } In this paper, the framework we proposed uses the backbone proposed by PanoFormer\cite{shen2022panoformer} as a common feature encoder. The backbone consists of the panorama transformer block and convolution 2D layer based downsample layer. The panorama transformer block contains a window self-attention mechanism designed for the panorama imaging process and a feed-forward layer designed based on depth-wise separable convolution\cite{Howard_2019_ICCV}. The entire backbone consists of an input projection layer and $4$﻿ stage blocks. The input projection layer consists of a convolution layer whose kernel size is $3 \times 3$﻿ with stride of $2$﻿, a 2D batch-norm layer and a ReLU activation function. The subsequent $4$ stage blocks all consist of a panorama transformer block and a double-downsampling layer composed of a 2D convolutional network. Each stage block will output the corresponding double-downsampled feature map result.

\textbf{Layout Decoder: } Considering that layout and camera pose estimation are non-pixel-level task while depth estimation and region segmentation are both pixel-level tasks. Some works\cite{zhang2021survey} mentioned that different tasks may have different requirements for features of different scales. Following the hohonet\cite{sun2021hohonet}, we first employ a height compression module to project multi-scale panoramic features from 2D into a 1D sequential representation. This sequence is then processed by the Transformer encoder introduced in Hohonet\cite{sun2021hohonet}, and finally passed through a lightweight MLP to predict the enclosed room layout $S_{room}$.

\textbf{Camera Pose Decoder:} The camera pose decoder follows a feature processing pipeline consistent with that of the layout decoder: multi-scale 2D panoramic features are first compressed along the height dimension into a 1D sequence, which is then processed by a Transformer encoder. However, due to the distinct nature of the camera pose estimation task, the corresponding feature processing blocks in this decoder do not share weights with those in the layout decoder. After feature encoding, the camera pose is regressed through a lightweight multi-layer perceptron (MLP) head.

\textbf{Depth Decoder: } In this section, we maintain consistency with the original PanoFormer\cite{shen2022panoformer} paper. This part of the network consists of panorama transformer blocks and linear interpolation upsampling layers. The decoder network is structurally symmetrical to the panorama encoder. On the last panorama transformer block's feature output $\hat{f}$﻿, the depth estimation decoder predict the coarse depth maps $S^{p}_{depth} = \{ d_{ij} | i=1,2,...,W; j=1,2,...,H \}$﻿.

\textbf{Region Segmentation Decoder:} Given that the panorama encoder and depth decoder together constitute a U-Net-like architecture—commonly employed in semantic segmentation—we do not introduce separate feature extraction modules for the region segmentation decoder. Instead, we directly utilize the feature map $\hat{f}$ output by the final transformer block in the depth decoder. Two independent convolutional layers are then applied to this feature map to predict the irregular region mask $S_{ir} = \{ p_{ij} \in {0, 1} | i =1, 2, ..., W; j=1, 2, ..., H \}$ and the background segmentation mask $S_{seg} = \{0 \leq p_{ij} \leq 1 | i = 1, 2, ..., W; j= 1, 2, ..., H \} $, respectively.

\subsection{Pose Aware Background Depth Resolving Component}
\begin{figure}[!t]
\centering
\includegraphics[width=3.5in]{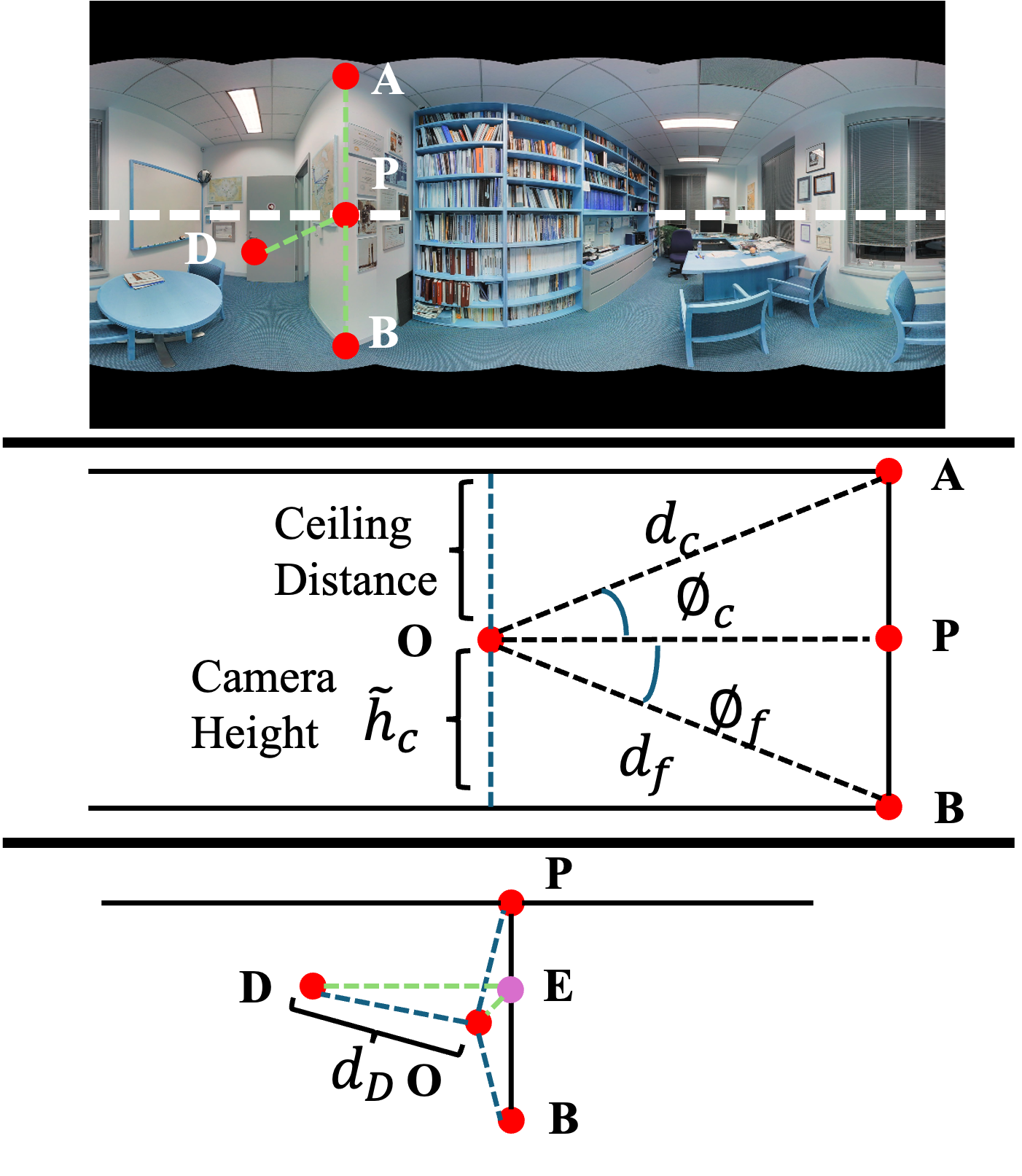}
\caption{P is the camera center, A and B are the upper and lower boundary points of the wall corresponding to point P, and D is an arbitrary point on the wall plane. The lengths of AB in the image are known, and the corresponding angles $\phi_{c}$ and $\phi_{f}$ can be calculated based on spherical camera geometry. Based on the depth of point P predicted by the depth decoder, $d_c$ and $d_f$ can be calculated.}
\end{figure}
\subsubsection{Camera Height Optimize}
Given a room layout estimation prediction $S_{room} \in R^{W \times 3}$ to resolve the corresponding background depth, we need to obtain the camera-to-ground and camera-to-ceiling distance information. Existing methods often assume a fixed camera height or camera height can be obtained easily. This assumption is stable and reliable in virtual rendering datasets, but it is not necessarily reliable when collecting data in the real world using a panoramic depth camera. To obtain a stable and reliable camera height, our framework first use a camera pose decoder to predict an init camera height $\hat{h}_c$ from input panorama. Then, the PA-DBA component use the depth, room layout decoder's prediction to calculate a camera height $\tilde{h}_c$﻿. Finally, we count the mean value of $\hat{h}_c$ and $\tilde{h}_c$﻿ to solve the accurate camera height $h_c$﻿. In the following, we show the detail of calculation process of $\tilde{h}_c$﻿.

From the results of room layout and region segmentation, we can obtain the foreground pixel coordinates of the upper and lower boundaries of the wall without occlusion. Here, we take a straightforward and easy-to-understand example (as the Fig. 3) to explain how to calculate our camera height $\tilde{h}_c$﻿ from these occlusion free wall boundaries pixel coordinates.

As shown in Fig. 3, we define the origin of the camera plane as $O$﻿, and then draw a horizon line on the camera plane centered at point $O$. Based on the results of the region segmentation decoder, we can select the portion of horizon line without occlusion and falling in the wall. We randomly select a point $P$ from these segment regions; the length of $OP$ is then the depth value of the corresponding pixel in the coarse depth map. Simultaneously, based on the horizontal coordinates of point $P$, we can obtain the two points $A$ and $B$ above and below the corresponding wall boundary line. The lengths of $OA$ and $OB$ can also be obtained from the corresponding depth map pixels $d_c$﻿ and $d_f$﻿. Using basic geometry, we can calculate the camera height $AP$ and the height $BP$ from the camera to the ceiling. The calculation formulas are as follows:
\begin{equation}
\begin{aligned}
|AP| = d_{c} \times sin(\phi_{c}), & \
|PB| = d_{f} \times sin{\phi_{f}}, &
\end{aligned}
\end{equation}
where angles $\phi_c$﻿ and $\phi_f$﻿﻿ can be calculated using the geometric imaging principle of a panoramic camera. Assuming that the vertical coordinate of point $A$﻿ is﻿ $u_{ceil}$﻿ and the vertical coordinate of point B﻿ is $u_{floor}$, the calculation formulas for the two angles can be obtained using the following formulas:

\begin{equation}
\begin{aligned}
\phi_{c} = (0.5 - \frac{u_{ceil}}{H}) \times \pi, &\\
\phi_{f} = (\frac{u_{floor}}{H} - 0.5) \times \pi, &
\end{aligned}
\end{equation}

Through the above process, we can obtain a set of camera heights and distances from the camera to the ceiling. However, this calculation process is greatly affected by depth estimation errors. To reduce this impact, we use the horizontal line passing through point $O$ as a reference and perform the above calculation process at each occlusion-free point $P$ on this line segment. Finally, we average all the calculated camera heights to obtain the final $\tilde{h}_c$. Ultimately, averaging $\hat{h}_c$ and $\tilde{h}_c$ yields a relatively accurate camera height $h_c$.

\subsubsection{Enclosed Background Depth Resolving}

After determining the camera height $h_c$ and the distance between the camera and the ceiling, we can use the wall information from the room layout predictions to calculate the corresponding background depth maps for each pixel belonging to the background wall. The depth maps for the ceiling and floor can be simply calculated based on the angles calculated from the pixels using the following formula:

\begin{equation}
\begin{aligned}
d^{i}_{floor} = \frac{h_c \times H}{(u^{i}_{floor} - 0.5H) \times \pi}, & \\
d^{i}_{ceil} = \frac{|AP| \times H}{(0.5H - u^{i}_{ceil}) \times \pi}, &
\end{aligned}
\end{equation}
 
For any point $D$ on the wall plane that is tilted toward the floor, assume its equirectangular plane coordinates on the panorama image are $(u_i,v_i)$. First, we can calculate the horizontal and vertical offset angles $\rho_D$ and $\phi_D$ of this point relative to the camera center $P$ using the following formulas: 

\begin{equation}
\begin{aligned}
\phi_{D} = (\frac{u_{i}}{H} - 0.5) \times \pi, & \\
\rho_{D} = (1 - \frac{v_i}{W}) \times \pi, &
\end{aligned}
\end{equation}

Based on the two deflection angles, to obtain the depth of $D$﻿, $d_{D} = |OD|$, we first draw a perpendicular line through point $D$﻿ to line segment $AB$﻿ and intersect $AB$﻿ at point $E$﻿. First, based on the vertical offset angle $\phi_{D}$, we can obtain $|OE|=cos(\phi_{D})|OP|$, where the value of $|OP|$ can be directly obtained from depth prediction. Then, based on the horizontal offset angle $\rho_{D}$﻿ and $|OE|$, we can simply solve $d_{D}=cos(\rho_{D})|OE|$. Similarly, any point on the wall plane that is deflected toward the ceiling can also use a similar method to solve for its corresponding depth value. Finally, combining all the solved depth values, we can obtain $S_{back}$.

\subsection{Fusion Mask Generation Component}
A crucial step in constraining the depth decoder's prediction using the calculated background depth is identifying which regions in the input panoramic image are valid. Previous work, BGDNet, simply used the semantic segmentation mask output by the SAM model as the criterion for this. This approach effectively distinguishes which parts of a room's interior are foreground objects and which areas belong to the background walls. However, this approach relies on two prerequisites: 1) no large areas in the input image extend beyond the room structure; and 2) the room structure itself is regular and conforms to the Manhattan layout. Previous work has mentioned that any irrationally shaped room structure can essentially be divided into regular and irregular regions. The room layouts typically provided by datasets describe enclosed regular regions within the room structure. Therefore, to make the background depth-based method more generalizable, we need to determine not only whether a pixel belongs to the foreground object but also whether the region belongs to the regularly enclosed region defined by the room layout.

In our framework, we treat the distinction between foreground and background regions, and between regular enclosed regions and irregular regions, as two independent binary semantic segmentation tasks. Both of these semantic segmentation tasks are performed by our region segmentation decoder. This decoder outputting the irregular region mask $S_{ir} $ and the background mask $S_{seg}$. In $S_{ir}$, each element $s^{ir}_{uv}$ represents the probability that the pixel at row $u$ and column $v$ in the input panorama belongs to an irregular region. Similarly, in $S^{seg}$, each element $s^{seg}_{uv}$ denotes the probability that the pixel belongs to the background. To properly fuse coarse depth prediction and background depth, we need to calculate a fusion weight to balance their respective proportions in the final output. Here, we first binarize each element in the region mask $S_{ir}$﻿ using a threshold of $0.5$. Then, we use the sigmoid function to compress the value of each element in $S_{seg}$﻿ to the range $[0,1]﻿$. Finally, we multiply each element in $S_{ir}$﻿ by each element in $S_{seg}﻿$ to obtain the final fusion weight mask $S_{weight}$.

\subsection{Adaptive Fusion Component}
As mentioned above, based on the depth decoder's prediction and the enclosed background depth map, our framework needs to fuse the predicted depth $S^{p}_{depth}$﻿ and the enclosed background depth $S_{back}$﻿ to obtain an accurate final depth estimation result $S^{final}_{depth}$﻿. Considering that $S_{back}$﻿ is naturally suitable as an upper bound for depth estimation when the layout prediction is relatively accurate.Then $S_{back}$﻿﻿ as an upper bound can be used to ensure that the depth of pixels originally belonging to the rescue, ceiling, ground and other areas belonging to regular regions do not exceed the area of the room itself. And, the depth value of the pixel belonging to the foreground object or irregular region can be as close to $S^{p}_{depth}$﻿﻿ as possible. The entire fusion can be described as following formula:

\begin{equation}
\begin{aligned}
S^{final}_{depth} & = \{d^{f}_{uv} | u = 1,2,...,W;v=1,2,...,H\};  & \\
etl. & \ d^{f}_{uv} = s^{uv}_{weight} \times d^{b}_{uv} + (1 - s^{uv}_{weight}) \times d^{p}_{uv},  \\
 & s^{uv}_{weight} \in S_{weight}; d^{b}_{uv} \in S_{back}; d^{p}_{uv} \in S^{p}_{depth},
\end{aligned}
\end{equation}

\subsection{Objective Function}
Our framework as a whole includes four task decoders: layout estimation, depth estimation, camera pose estimation, region segmentation. The objective functions corresponding to the four decoders can be expressed as $L_{layout}$﻿, $L^{p}_{depth}$﻿, $L_{region}$, and $L_{pose}$, respectively. 
Since most existing datasets lack comprehensive annotations for room layout estimation, we first aggregate all layout-related annotations from the Structured3D and Matterport3D datasets to form a large-scale layout dataset. Following the practice described in\cite{10657716}, we apply a simple script to convert the existing room layout annotations—which may contain occlusions or ambiguities—into regular and enclosed layouts. Using this generated dataset, we first pre-train the room layout branch of our proposed framework. For this stage, similar to HorizonNet, we adopt a combination of binary cross-entropy loss and L1 loss. For the depth estimation branch, we follow Panoformer and employ a Huber\cite{esmaeili2019novel} loss together with a gradient loss\cite{shen2022distortion}, keeping the hyper-parameter settings consistent with those in the original HorizonNet\cite{sun2019horizonnet} and Panoformer papers respectively. For the two semantic segmentation subtasks, we use the commonly adopted binary cross-entropy loss and Dice loss\cite{li-etal-2020-dice}. As for the camera pose estimation, we simply apply a smooth L1 loss. Overall, the total loss function of our proposed framework can be expressed as:
\begin{equation}
L_{all} = \lambda_1 * L_{layout} + \lambda_2 * L_{depth} + \lambda_3 * L_{region} + \lambda_4 * L_{pose},
\end{equation}
where $\lambda_1, \lambda_2, \lambda_3, \lambda_4$ are four manual setting parameters.

\section{Experiment}
\subsection{Experiment Setting}
In this section, we introduce the experiments to verify the proposed depth estimation framework. In our experiments, we selected three datasets: MatterPort3D\cite{chang2017matterport3d}, Structure3D\cite{zheng2020structured3d}, Replica\cite{szot2021habitat}. MatterPort3D and Reiplica is collected in the real world, while the Structure3D dataset is a synthesized dataset. 

\textbf{Matterport3D:} Matterport3D contains 10,800 panoramic images collected from 90 different rooms. The camera used in the collection process of this dataset is Matterport's Pro 3D camera. In this part of the dataset, we also use 61 room images for training and 29 room images for testing. All RGB and depth images are also downsampled to $512\times1024$ size.

\textbf{Structured3D:} The Structured3D dataset contains 196K rendered panoramic images and corresponding depth maps, covering 12,835 rooms in 3,500 scenes. Each room is created manually using CAD models of furniture, which are in real-world dimensions and used in real production. In this dataset, we follow the official recommended setting \cite{zheng2020structured3d}, using the data of the first 3000 scenes as the training set, the data of 3000-3250 scenes as the validation set, and the data of the last 250 scenes as the test set.

\textbf{Replica:} The Replica dataset is captured using an RGB-D camera rig with an IR projector from real-world indoor scenes. We follow the BGDNet\cite{chen2024bgdnet} render 3554 images from these 3D models by setting camera in random position on the floor.

\textbf{Metric:} Following the previous works\cite{bai2024glpanodepth, shen2022panoformer, zhang2025sgformer}, we use some standard evaluation metrics to measure our methods, which include: mean relative error (MRE), mean average error (MAE), root mean squared error (RMSE), and three threshold percentage $\delta < \varsigma^t (\varsigma = 1.25, t=1,2,3)$ denoted as $\delta^t$.  For the room layout task, we used 2D IoU, 3D IoU, RMSE, and $\delta^1$ for evaluation. For the camera pose task, we used relative rotation error(RRA), relative translation error(RTA), and area under the curve(AUC) for evaluation, with a threshold of 30 for each metric. For the region segmentation task, we used mIoU for evaluation.

\textbf{Training setting: } We use Adam as the optimizer, and the parameters of the optimizer are basically the basic settings of the pytorch framework. For the learning rate scheduling strategy, we choose one-cycle\cite{smith2019super}, set the initial learning rate to 0.0001, and the minimum learning rate to 0.0000001. Our hardware experimental platform is configured with AMD Epyc 7003 CPU and 4-card RTX 4090 GPU. During the training process, we set the batch size on each card to 2. In the data set enhancement part, we used random horizontal angle rotation and random horizontal flipping with reference to Panoformer\cite{shen2022panoformer}. For horizontal angle rotation, we set the interval of random angle to $[-\frac{\pi}{4}, \frac{\pi}{4}]$.

\textbf{Parameter setting: }The training objective function used by our framework is shown in Section \uppercase\expandafter{\romannumeral 3}. B , which contains four subcomponents: $L_{pose}$, $L_{region}$, $L_{depth}$, and $L_{layout}$. During pre-training of the layout decoder, we set the $\lambda_1$ to 0 and other parts to 0, while during the training the training of the other decoders, we set $\lambda_2, \lambda_3, \lambda_4$ as 1.0, 0.4, 0.6.  Other hyper-parameters that are not relevant to this study are set according to the common specifications in the current field.

\begin{table*}[!t]
\centering
\caption{Quantification comparison with State of the Art depth depth estimation methods on Matterport3D}
\begin{tabular}{lllcccccc}
\toprule[1.2pt]
\multicolumn{1}{c}{\multirow{2}{*}{Dataset}} & \multicolumn{1}{l}{\multirow{2}{*}{Method}} & \multicolumn{1}{l}{\multirow{2}{*}{Pub' Year}} & \multicolumn{6}{c}{Classic Metrics}                                                                             \\ 
\multicolumn{1}{c}{} & \multicolumn{1}{l}{} & \multicolumn{1}{l}{} & $\delta_1$ & $\delta_2$ & $\delta_3$ & RMSE       & MRE        & MAE        \\ 
\midrule[1.2pt]
\multirow{10}{*}{Matterport3D}    & Panoformer\cite{shen2022panoformer} & ECCV 2022 & 0.9184 & 0.9804  & 0.9916  & 0.3635 & 0.0571 & 0.1013     \\
   & EGFormer\cite{yun2023egformer} & ICCV 2023 & 0.8158 & 0.9390 & 0.9735 & 0.6025 & 0.1517  & 0.1473     \\
   & HRDFuse\cite{ai2023hrdfuse} & CVPR 2023 & 0.9162  & 0.9669  & 0.9844 & 0.4433  & 0.0936 & 0.0967     \\
   & GLPanoDepth\cite{bai2024glpanodepth} & IEEE TIP 2024 & 0.8641 & 0.9561 & 0.9808 & 0.5223 & - & 0.2998 \\
   & GADFNet\cite{huang2025gadfnet} & IEEE TCSVT 2025 & 0.9047 & 0.9668 & 0.9868 & 0.4564 & 0.0963 & 0.2592 \\
   & SN360\cite{11084791} & IEEE Access 2025 & 0.4483 & - & - & 0.9392 & 0.9808 & 0.9932 \\
   & SGFormer\cite{zhang2025sgformer} & IEEE TCSVT 2025 & 0.8946 & 0.9642 & 0.9859 & 0.4790 & - & 0.2748 \\
   & DepthAnyPanorama\cite{lin2025depth} & arxiv 2025 & 0.8518 & 0.9113 & 0.9764 & 0.7510 & 0.4456 & 0.7845 \\
   & DepthAnyDirection\cite{li2025depth}  & ICLR 2026 & \textbf{0.9561} & \textbf{0.9860} & 0.9982 & 0.2882 & 0.0766 & 0.0912 \\
   & \textbf{Ours}  & & 0.9399 & 0.9852 & \textbf{0.9984} & \textbf{0.2236} & \textbf{0.0411} & \textbf{0.0629} \\
\bottomrule[1.2pt]
\end{tabular}
\end{table*}

\begin{table*}[!t]
\centering
\caption{Quantification comparison with State of the Art depth estimation methods on Structured3D}
\begin{tabular}{lllcccccc}
\toprule[1.2pt]
\multicolumn{1}{c}{\multirow{2}{*}{Dataset}} & \multicolumn{1}{l}{\multirow{2}{*}{Method}} & \multicolumn{1}{l}{\multirow{2}{*}{Pub' Year}}  & \multicolumn{6}{c}{Classic Metrics}                                                                             \\ 
\multicolumn{1}{c}{}                         & \multicolumn{1}{l}{}   & \multicolumn{1}{l}{} & $\delta_1$ & $\delta_2$ & $\delta_3$ & RMSE       & MRE        & MAE        \\  
\midrule[1.2pt]
\multirow{9}{*}{Structure3d} & PanoFormer\cite{shen2022panoformer} & ECCV 2022 & 0.8943 & 0.9536 & 0.97431 & 0.3017 & 0.1201 & 0.1546 \\   
					   & EGFormer\cite{yun2023egformer}  & ICCV 2023  & 0.7979 & 0.9071 & 0.9455 & 0.6841 & 0.4509 & 0.2205     \\
                                             & HRDFuse\cite{ai2023hrdfuse} & CVPR 2023 & 0.7561 & 0.9161 & 0.9631 & 0.4061 & -  & 0.2451     \\
                                             & SGFormer\cite{zhang2025sgformer} & IEEE TCSVT 2025 & 0.9613 & 0.9896 & 0.9957 & 0.2429 & - & - \\
                                             & GADFNet\cite{huang2025gadfnet} & IEEE TCSVT 2025 & 0.8647 & 0.9468 & 0.9899 & 0.3524 & 0.0812 & 0.1944 \\
                                             & DepthAnyPanorama\cite{lin2025depth} & arxiv 2025 & 0.8918 & 0.9243 & 0.9813 & 0.6641 & 0.4456 & 0.5125 \\   
                                             & DepthAnyDirection\cite{li2025depth} & ICLR 2026 & 0.9361 & 0.9713 & 0.9912 & 0.2914 & 0.0686 & 0.0992 \\  
                                             & \textbf{BGDNet\cite{chen2024bgdnet}}&  CVPR 2024 & 0.8336 & 0.9377 & 0.9731 & 0.3490 & -  & 0.1656     \\    
                                             & \textbf{Ours} & & \textbf{0.9679} & \textbf{0.9907} & \textbf{0.9983} & \textbf{0.1935} & \textbf{0.0414} & \textbf{0.0613} \\ \bottomrule[1.2pt]
\end{tabular}
\end{table*}

\begin{table*}[!t]
\centering
\caption{Quantification comparison with State of the Art depth estimation methods on Replica}
\begin{tabular}{lllcccccc}
\toprule[1.2pt]
\multicolumn{1}{c}{\multirow{2}{*}{Dataset}} & \multicolumn{1}{l}{\multirow{2}{*}{Method}} & \multicolumn{1}{l}{\multirow{2}{*}{Pub' Year}}  & \multicolumn{6}{c}{Classic Metrics}                                                                             \\ 
\multicolumn{1}{c}{}                         & \multicolumn{1}{l}{}   & \multicolumn{1}{l}{} & $\delta_1$ & $\delta_2$ & $\delta_3$ & RMSE       & MRE        & MAE        \\  
\midrule[1.2pt]
\multirow{5}{*}{Replica} & HRDFuse\cite{ai2023hrdfuse}  & CVPR 2023 & 0.7711 & 0.9221 & 0.9541 & 0.3891 & - & 0.1421 \\
				    & DepthAnyPanorama\cite{lin2025depth} & arxiv 2025 & 0.9012 & 0.9335 & 0.9898 & 0.4432 & 0.2121 & 0.3132 \\   
                                      & DepthAnyDirection\cite{li2025depth} & ICLR 2026 & 0.9431 & 0.9668 & 0.9933 & 0.2712 & 0.0733 & 0.1011 \\  
				    & \textbf{BGDNet}\cite{chen2024bgdnet} & CVPR 2024 & 0.8554 & 0.9365 & 0.9624 & 0.3456 & - & 0.1678 \\
				    & \textbf{Ours} & & \textbf{0.9233} & \textbf{0.9866} & \textbf{0.9992} & \textbf{0.2101} & \textbf{0.0512} & \textbf{0.0629} \\
				    
\bottomrule[1.2pt]
\end{tabular}
\end{table*}

\subsection{Region Segmentation Decoder's Annotation Generation}
\subsubsection{The generation process of background segmentation}

To efficiently fuse the predictions from the background and depth decoders, we first need information to guide the model in identifying which regions belong to the background and which to the foreground. For region segmentation, we refer to BGDNet's approach, using the corresponding semantic segmentation results to uniformly set all ground, ceiling, and wall surfaces in the scene as the background region. Unlike BGDNet, which uses SAM to segment the input image, we directly use the semantic segmentation annotations provided by the dataset to generate the corresponding masks.

\subsubsection{The generation process of irregular region segmentation}

As mentioned in the main text, the background depth generated by BGDNet often ignores the fact that in some scenes, the depth threshold may exceed the background space defined by the room layout, or the entire indoor scene itself may not be a regular Manhattan layout. However, non-Manhattan layouts or irregularly shaped rooms will always intrude a regular enclosed Manhattan region, and the room layout annotations provided in the dataset are usually this regular enclosed Manhattan region. Therefore, to make this approach of fusing background and depth decoders more generalizable, we need information to help the model distinguish which regions belong to the regular enclosed region.

Here, to generate a reasonable irregular region segmentation mask, we first calculate the scene background depth for each panel input image using the room layout annotations and camera height from the entire dataset. Then, we use the generated background segmentation mask annotation to select the depth of the fused region from both the background depth and ground truth depth annotations. Finally, we calculate the difference between the depths of the fused regions selected from the two depths and apply a threshold to obtain the corresponding irregular region segmentation mask annotation. In the process of creating the binary mask, we use a threshold of 0.05, which is based on our specific observations in the experiments.

\subsection{Performance Comparison}

\subsubsection{Quantification Comparison with SoTA Depth Estimation Methods}

\begin{figure*}[!t]
\centering
\includegraphics[width=6.7in]{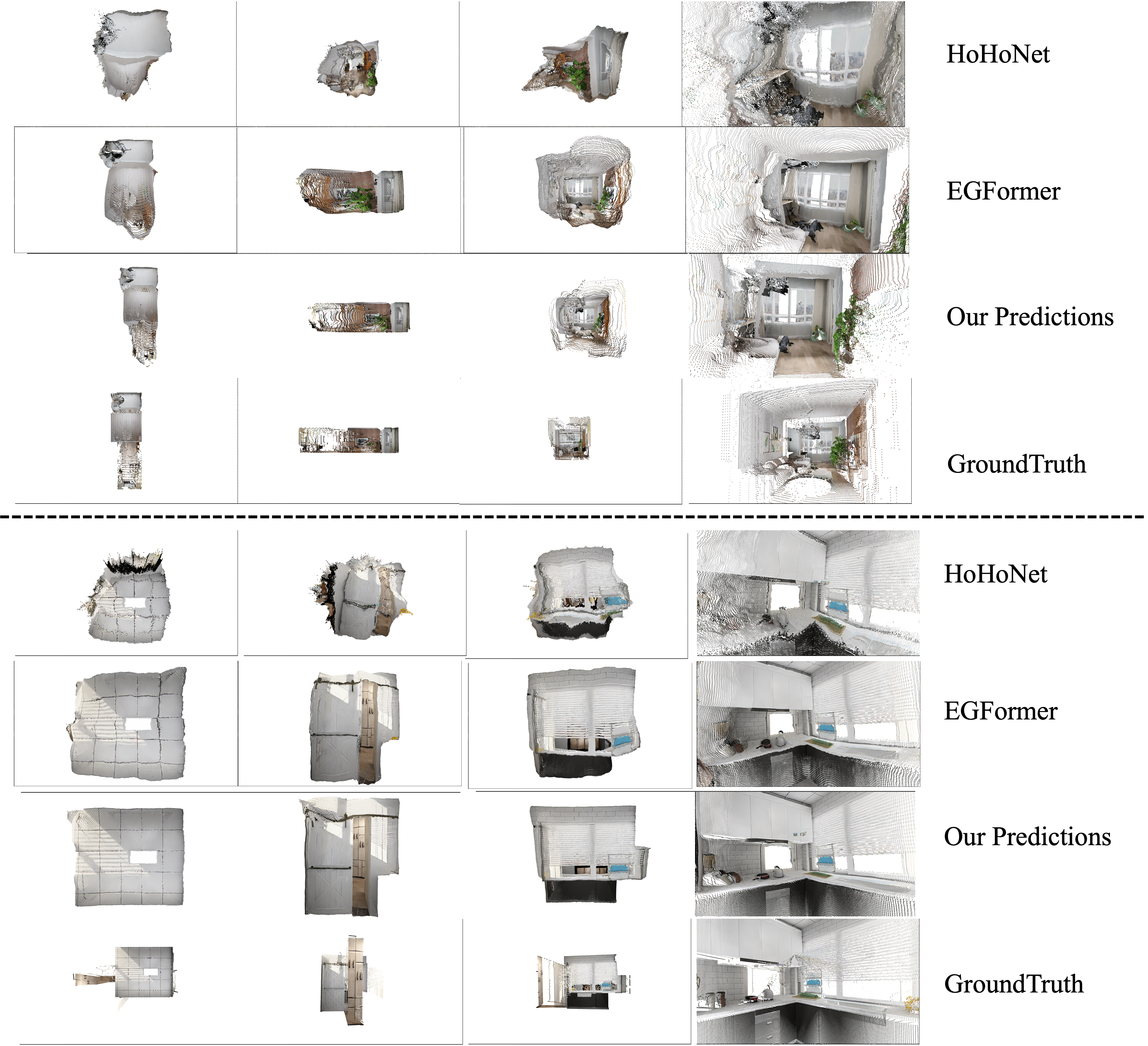}
\caption{The 3D visualize results of final depth estimation. For each scene, we selected three perspectives: top view, side view, front view, and internal perspective to display the three-dimensional visualization effect of the point cloud converted from the depth map.}
\setlength{\abovecaptionskip}{-0.3cm}
\end{figure*}

In this section, we will conduct performance comparison experiments on three datasets: Matterport3d, Structured3d, and Replica. The results have been shown in Table  \uppercase\expandafter{\romannumeral 1}- \uppercase\expandafter{\romannumeral 3} . To validate the effectiveness of our approach, we compare our RGCNet with the current state-of-the-art methods\cite{yun2023egformer, li2022omnifusion, wang2020bifuse, jiang2021unifuse, ai2023hrdfuse, bai2024glpanodepth, shen2022panoformer, zhang2025sgformer, chen2024bgdnet, berenguel2022fredsnet, He_2022_CVPR, 11084791}, including strategies for bi-projection fusion, long-range dependencies, multi-task learning, and background-based methods. 

On the Matterport3D dataset, our method also achieved the current best performance in RMSE, the core evaluation indicator of depth estimation, which fully demonstrated the effectiveness of the proposed method and data processing strategy in key accuracy metrics. However, on the relative error threshold indicator ($\delta <$  1.25), our method failed to achieve the best results. We believe this is related to the introduced layout constraints: this constraint tends to optimize the absolute error between the prediction and the true value, but in areas with small true depth values, this optimization may result in a limited range of change in the predicted value, making it more difficult to meet the strict 1.25 times relative error requirement. Despite this, we still achieved a level close to SOTA on this indicator. Importantly, this method is significantly ahead of existing work in RMSE.

On the Structured3d dataset, we mainly compare with BGDNet, which uses a similar idea. The results released in the BGDNet paper are trained on the replica and then val on the structure3d dataset, while our method is trained on Structured3D and then validated on the corresponding validation set. Due to the differences in the settings of train and val, the method proposed in this paper has a very obvious advantage in the six indicators of concern.

On the Replica dataset, we ensured that all data processing steps were as consistent as possible with BGDNet. Based on the performance across the six metrics shown, our method significantly outperforms BGDNet. Furthermore, it demonstrates a clear advantage over the current state-of-the-art methods, DepthAnyPanorama and DepthAnyDirection. It's worth noting that, considering both of these methods used large datasets to pre-train large models, the performance we present is the result of fine-tuning the officially provided weights on the Replica dataset for two epochs.

Overall, in terms of quantitative metrics, our method demonstrates a significant performance advantage compared to current state-of-the-art methods, especially in the most common metric rmse. This clearly shows that using the background depth we constructed effectively constrains the depth estimator's predictions, leading to more accurate depth estimation results.

\subsubsection{Visualize Comparison with SoTA Depth Estimation Method}

In this section, we provide a qualitative evaluation of depth estimation results, with visual comparisons shown in Fig. 4. The selected comparison methods include HoHoNet\cite{sun2021hohonet}, a multi-task learning approach, and EGFormer\cite{yun2023egformer}, which is based on a panoramic Transformer architecture. The visual results demonstrate that, owing to the effective constraint provided by our background depth modeling, our framework captures the overall geometric structure of the room more accurately than existing open-source state-of-the-art methods. This improvement also validates the accuracy of the enclosed-region background depth recovered by our pose-aware depth resolving strategy.

It should be noted that while our framework predicts the overall room geometry more accurately, its performance remains limited in areas outside the enclosed background region. Specifically, in the first scene, distant objects such as the sofa are not well reconstructed. In the second scene, certain protruding structures are incorrectly compressed onto the wall planes in the depth prediction.

The limitation observed in the first scenario is primarily attributed to the fact that the current enclosed-region layout estimation can only represent a limited portion of the overall room structure. For depth prediction in areas beyond the enclosed region, our framework relies principally on the capacity of the base depth decoder. Nevertheless, accurate depth estimation in such areas remains a persistent challenge for panoramic depth estimators in indoor scenes, and a widely effective solution has yet to be established.

The issue observed in the second scenario can be attributed to current limitations in irregular region segmentation. Inherent inaccuracies in this task lead to certain areas outside the enclosed background being misclassified as background regions. Consequently, these areas—which should reflect the room’s extended structure—are incorrectly flattened onto the wall planes in the depth prediction.

\subsection{Quantification Analysis of Background Depth Resolving}
\begin{table}[!t]
\caption{The performance of our framework's layout decoder}
\centering
\begin{tabular}{lllll}
\toprule[1.2pt]
Methods & 2DIoU & 3DIoU & RMSE & $\delta^1$ \\ \midrule[1.2pt]
HorizonNet\cite{sun2019horizonnet} & 91.18 & 89.77 & 0.09 & 0.99 \\
HoHoNet\cite{sun2021hohonet} & 92.96 & 91.52 & 0.09 & 0.99 \\
Ours & 94.36 & 92.96 & 0.04 & 0.99 \\
\bottomrule[1.2pt]
\end{tabular}
\end{table}

\begin{table}[!t]
\caption{The accuracy of our framework's pose decoder}
\centering
\begin{tabular}{llll}
\toprule[1.2pt]
Methods & RRA@30 & RTA@30 & AUC@30 \\ \midrule[1.2pt]
Ours & 94.33 & 90.51 & 91.34 \\ \bottomrule[1.2pt]
\end{tabular}
\end{table}

\begin{table}[!t]
\caption{The accuracy of camera height}
\centering
\begin{tabular}{lc}
\toprule[1.2pt]
 & mean average error \\ \midrule[1.2pt]
camera height from pose decoder $\tilde{h}_c$ & 0.05 \\
camera height from depth decoder $\hat{h}_c$ & 0.04 \\
camera height from resolving strategy $h_c$ & 0.02 \\
 \bottomrule[1.2pt]
\end{tabular}
\end{table}

\begin{table}[!t]
\centering
\caption{The accuracy of resolved background depth maps.}
\begin{tabular}{llll}
\toprule[1.2pt]
Methods       & RMSE   & MRE    & MAE    \\ \midrule[1.2pt]
HoHoNet's Background   & 0.3514 & 0.1223 & 0.1923 \\
PanoFormer's Background & 0.2644 & 0.0688 & 0.0991 \\
Layout GT with Fix Camera pose & 0.5514. & 0.3846 & 0.3102 \\
Layout GT with GT Camera pose & 0.0011 & 0.0010 & 0.0010 \\
Layout GT with Pred Camera pose & 0.0013 & 0.0012 & 0.0012 \\ 
Layout Pred with Fix Camera pose & 0.5862 & 0.3212 & 0.3833 \\
Layout Pred with GT Camera pose & 0.0403 & 0.0312 & 0.0213 \\ 
Layout Pred with Pred Camera pose & 0.0551 & 0.0417 & 0.0412 \\ \bottomrule[1.2pt]
\end{tabular}
\end{table}
In this section, we will construct corresponding experiments to verify and compare the performance differences between our background depth calculation strategy and BGDNet's background depth calculation strategy. For convenience, we will only construct experiments on the Structured3D dataset. We will first demonstrate the overall performance of the framework's layout decoder and camera pose parts. Based on this, we will show the RMSE, MRE, and MAE of the background region depth and depth ground truth calculated by layout ground truth and layout prediction with and without camera pose. Furthermore, we will demonstrate the corresponding metrics of the background depth calculated using our method based on camera pose.

The results of layout estimation and camera pose estimation in the framework are shown in Tables \uppercase\expandafter{\romannumeral 4} and  \uppercase\expandafter{\romannumeral 5}, respectively. Table  \uppercase\expandafter{\romannumeral 6} compares the camera height $\hat{h}_c$ calculated from the depth decoder, the camera height $\tilde{h}_c$ obtained from the camera pose decoder, and the average of the two $h_c$ in our pose-aware background depth resolving strategy. The results in Tables \uppercase\expandafter{\romannumeral 4} and \uppercase\expandafter{\romannumeral 5} show that the layout decoder and pose estimation results in our framework are quite accurate. However, the results in Table 6 show that the camera height calculated from either the camera pose decoder or the depth decoder still has an error of about 5 cm; however, the error of the camera height obtained by averaging the two is reduced to 2 cm. This fully demonstrates the effectiveness and rationality of the pose estimation strategy in our proposed method.

Table \uppercase\expandafter{\romannumeral 7} presents the accuracy of our baseline model in background depth and the experimental results related to background depth calculated based on layout. The results show that the depth prediction results of our chosen open-source baseline model are relatively poor in the background portion. Furthermore, the background depth value calculated by room layout is highly dependent on the camera height. The results in rows 3 and 6 indicate that the camera pose used in generating images in different virtual scenes varies in the structured3d dataset. Rows 4 and 7 show that the background depth generated by ground truth layout can be considered, to some extent, as ground truth depth. The experimental results in rows 5 and 8 show that the background depth obtained by our proposed background depth calculation strategy also has some minor errors, but these errors are still orders of magnitude different from those produced by other current depth estimators.

\subsection{Quantification and Visualize Analysis of Fusion Mask Generation Component}
\begin{table}[!t]
\centering
\caption{The mIoU of background segmentation of our region segmentation decoder}
\begin{tabular}{lc}
\toprule[1.2pt]
Methods & mIoU \\ \midrule[1.2pt]
HoHoNet\cite{sun2021hohonet} & 72.12 \\
SAN\cite{xu2023side} & 95.33 \\
Ours & 89.23 \\
\bottomrule[1.2pt]
\end{tabular}
\end{table}

\begin{table}[!t]
\centering
\caption{The mIoU of irregular region segmentation of our region segmentation decoder}
\begin{tabular}{lc}
\toprule[1.2pt]
Methods & mIoU \\ \midrule[1.2pt]
HoHoNet\cite{sun2021hohonet}  & 73.33 \\
SAN\cite{xu2023side} & 96.32 \\
Ours & 90.92 \\
\bottomrule[1.2pt]
\end{tabular}
\end{table}

\begin{figure*}[!t]
\centering
\includegraphics[width=7in]{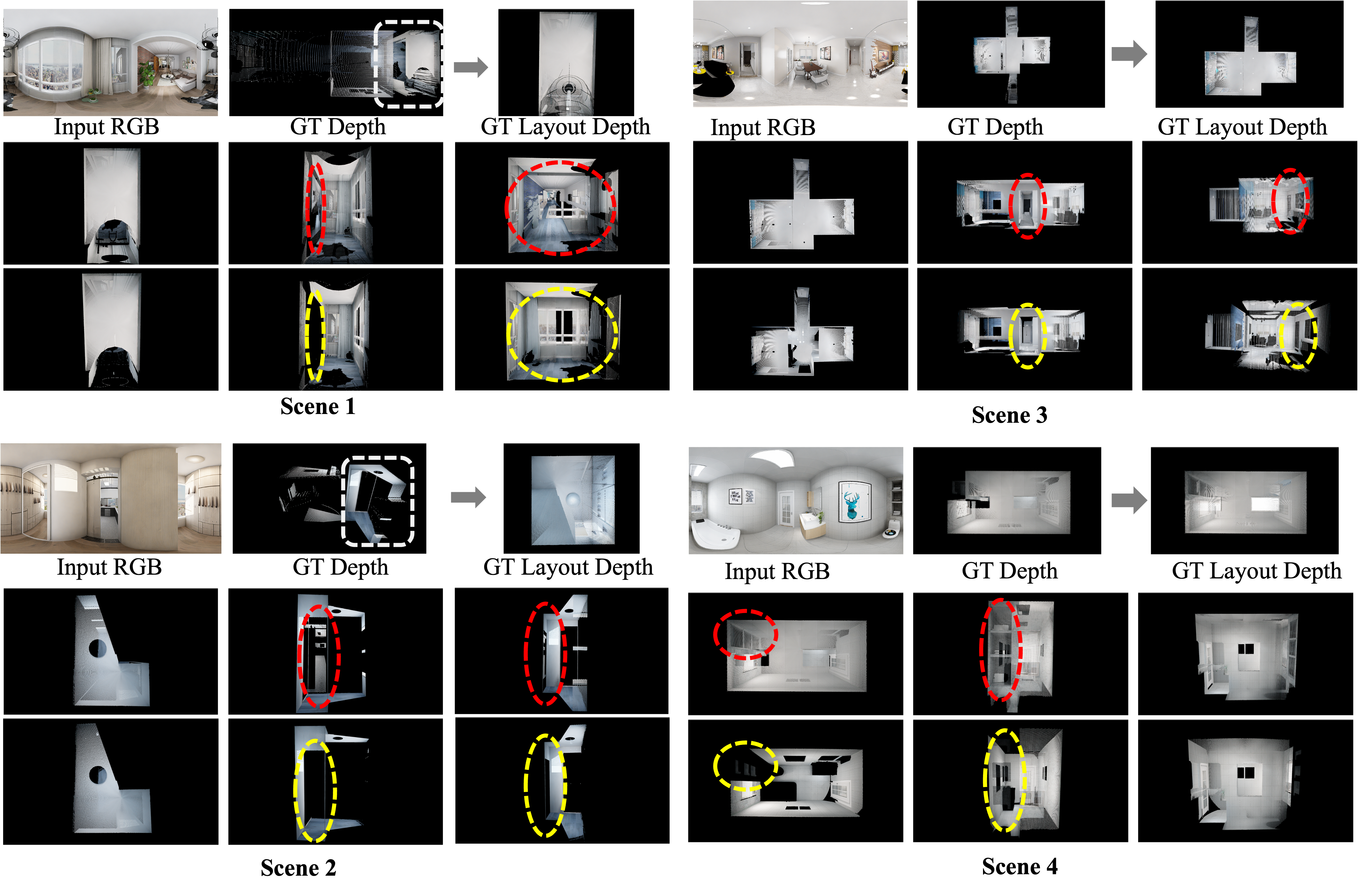}
\caption{The visualize of resolved background depth. The figure shows 3D visualizations of the background depth in four scenes constructed using our method and BGDNet. The areas marked with red circles represent regions that BGDNet misjudged as exceeding the background depth, while the areas marked with yellow circles represent regions that our constructed background depth accurately identifies.}
\end{figure*}

In this section, we construct experiments for the region segmentation decoder and the Fusion Mask Generation Component and verify their performance. The dataset used in the experiment is the same as that in the previous subsection, using Structured3D. Tables \uppercase\expandafter{\romannumeral 8} and \uppercase\expandafter{\romannumeral 9} show the performance of the two semantic segmentations performed by the region segmentation decoder. In these two subtasks, we chose the open-source HoHoNet as the comparison object. We did not choose newer methods because recent methods using multi-task learning on panoramic images are not open-source, so we could not conduct corresponding experiments on our own constructed experimental setup. Regarding our Fusion Mask Generation Component, the visualization results of our constructed experiments are shown in Fig. 5.

The experimental results in the two tables show that HoHonet exhibits significantly higher performance than the original paper due to the substantial reduction in task difficulty. However, considering HoHonet's overly aggressive approach of compressing 2D features of panoramic images to 1D, its performance still lags considerably behind our method. It's worth noting that while our method represents a significant performance improvement over earlier multi-task learning methods on panorama, it still lags behind recent methods specifically designed for semantic segmentation tasks, such as SAN\cite{xu2023side}. We believe this difference largely stems from our reuse of a portion of the depth decoder network for model overhead considerations.

As shown in Fig. 5, BGDNet does not consider that the depth of some regions in panoramic images in real-world scenes may exceed the space defined by the room layout when constructing the background depth. This leads to a disastrous problem for BGDNet when constructing the background depth: regions that should extend beyond the background depth are compressed onto the walls defined by the room layout. Although the binary semantic segmentation mask provided by SAM used by BGDNet can effectively distinguish which regions are walls, floors, and ceilings, the semantic segmentation mask provided by SAM cannot help the model identify which regions extend beyond the room layout, and these regions also belong to walls, floors, or ceilings. In contrast, our method benefits from the irregular region mask provided by the region segmentation decoder, which can effectively distinguish which parts of the scene extend beyond the room layout. It is worth noting that since the irregular region mask and background mask themselves are not completely accurate, the background depth constructed by our method has fewer regions for fusion, which reduces the constraint effect of the background depth to some extent. However, we believe that reducing some of the blended areas to make the background depth itself more generalizable is a reasonable trade-off.

\subsection{Quantification Analysis of Adaptive Fusion Component}
\begin{table}[!t]
\centering
\caption{The comparison of our proposed fusion methods and BGDNet}
\begin{tabular}{llll}
\toprule[1.2pt]
Methods       & RMSE   & MRE    & MAE    \\ \hline
Baseline      & 0.3017 & 0.1201 & 0.1546 \\
Fusion BGDNet\cite{chen2024bgdnet} & 0.2643 & 0.0812 & 0.1122 \\
Fusion ours   & 0.1953 & 0.0414 & 0.0613 \\  \bottomrule[1.2pt]
\end{tabular}
\end{table}
In this section, we construct corresponding experiments to verify the performance of our Adaptive Fusion Component. The relevant experimental results are shown in Table \uppercase\expandafter{\romannumeral 10}. The dataset used in the experiments remains structured3d. It is worth noting that, to ensure a fair comparison, we also used the irregular region mask provided by the region segmentation decoder of our framework in the experiments against BGDNet to ensure the accuracy of its background depth. As can be seen from the experimental results in Table \uppercase\expandafter{\romannumeral 10}, our fusion strategy is significantly better than BGDNet's approach. From the perspective of model structure, the region segmentation decoder in our framework is functionally equivalent to the module in BGDNet that extracts and fuses background depth features, while the subsequent threshold replacement operation in BGDNet functionally corresponds to our adaptive fusion component. Based on these analyses, we speculate that the advantage of our fusion method stems from our explicit learning of the weight process for distinguishing foreground and background regions as a supervised learning task during training. This learning-based approach is superior to implicitly fusing information into the feature extraction network.

\subsection{Ablation Study}

\begin{table}[!t]
\centering
\caption{The training impact of each task decoder to depth decoder}
\begin{tabular}{llllll}
\toprule[1.2pt]
\begin{tabular}[c]{@{}l@{}}
Depth \\ Decoder\end{tabular} & \begin{tabular}[c]{@{}l@{}}Camera Pose \\ Decoder\end{tabular} & \begin{tabular}[c]{@{}l@{}}Region \\ Decoder\end{tabular} & RMSE   & MRE    & MAE    \\ \midrule[1.2pt]
\checkmark &   &  & 0.3017 & 0.1201 & 0.1546 \\
\checkmark  & \checkmark & & 0.3011 & 0.1109 & 0.1411 \\
\checkmark &  &\checkmark & 0.2993 & 0.1182 & 0.1424 \\
\checkmark  &\checkmark &\checkmark & 0.3012 & 0.1192 & 0.1504 \\ \bottomrule[1.2pt]
\end{tabular}
\end{table}

\begin{table}[!t]
\centering
\caption{The training impact of each component to the final depth predictions.}
\begin{tabular}{llllll}
\toprule[1.2pt]
\begin{tabular}[c]{@{}l@{}}
Background \\ Resolving\end{tabular} & \begin{tabular}[c]{@{}l@{}}Fusion Mask \\ Generation\end{tabular} &\begin{tabular}[c]{@{}l@{}} Adaptive \\ Fusion \end{tabular} & RMSE & MRE & MAE \\ \midrule[1.2pt]
 & & & 0.3017 & 0.1201 & 0.1546 \\
 \checkmark & & & 0.2843 & 0.1011 & 0.1322 \\
 \checkmark & \checkmark & & 0.2231 & 0.0672 & 0.0833 \\
 \checkmark & \checkmark & \checkmark & 0.1935 & 0.0414 & 0.0613 \\
 \bottomrule[1.2pt]
\end{tabular}
\end{table}

\begin{figure}[!t]
\centering
\includegraphics[width=3.4in]{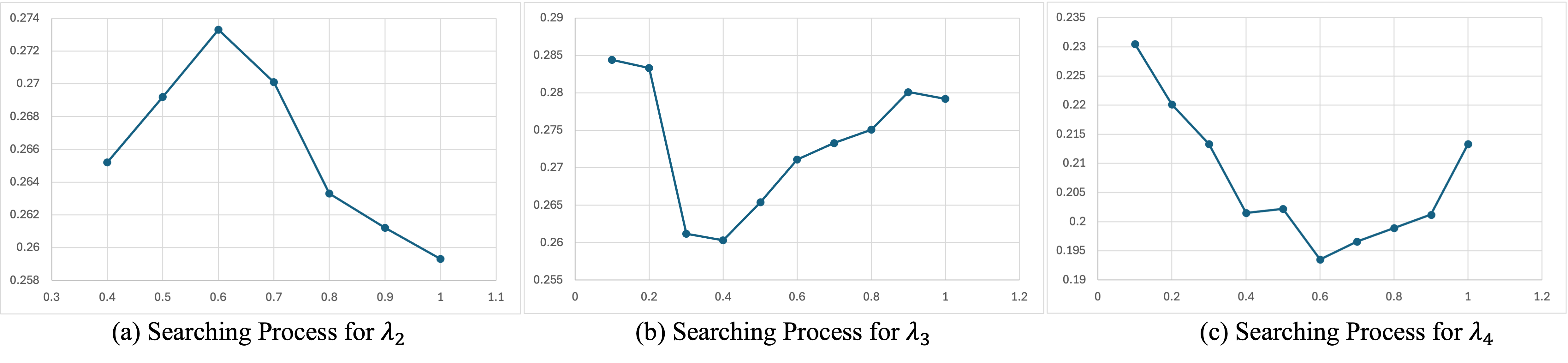}
\caption{The experiment results of explore the inference of hyper-parameter $\lambda_2$, $\lambda_3$ and $\lambda_4$. The x-axis in this figure represents the value of the corresponding hyper-parameter setting, while y-axis represents the RMSE value of our proposed framework's depth prediction.}
\setlength{\abovecaptionskip}{-0.3cm}
\end{figure}

In this section, we analyze the impact of each proposed component on the whole. Considering that we proposed a framework for auxiliary-task assisted depth estimation tasks, we mainly discuss two aspects of the impact of each component: 1) The impact of hyper-parameter setting for final depth predictions ;2) The impact of auxiliary-task decoder for depth estimation decoder;  3) The impact of each component for the final depth predictions. The experimental dataset we chose is Structured3D because it contains complete layout estimation annotations.

Regarding aspect 1) which we want to discuss, the relevant experimental results are shown in Fig. 6. In this set experiments, we divided the weight values of each component in the objective function within the range 0-1 into 0-1 into intervals of 0.1. We adopted a component-by-component exploration mode for configuring the optimal loss weights. Each time we explored the optimal configuration for a parameter, we set all subsequent parameters to 0 while keeping the preceding parameters unchanged. Considering that the layout decoder and other task decoders are trained independently in our framework, the value of its corresponding parameter $\lambda_1$ is constant at 0 when training other decoders. As shown in Fig. 6, the framework achieves optimal overall performance when $\lambda_2$, $\lambda_3$, and $\lambda_4$ are set to 1.0, 0.4, and 0.6, respectively.

Regarding aspect 2), the relevant experimental results are shown in Table  \uppercase\expandafter{\romannumeral 11}. In this set of experiments, we simply discuss the impact of each task decoder involved in the training process on the depth decoder results. In this set of experiments, the Loss weights for each task decoder used the optimal configuration explored in the previous set of experiments. As the results shown in the table, different task decoders do not have a significant impact on the final result during training in our framework. This also suggests that the parameters explored in the previous set of experiments actually have a greater impact on the final experimental results than on the performance of each task decoder, which in turn affects the performance of the three components we proposed, ultimately impacting the overall framework performance.

Regarding aspect 3), the relevant experimental results are shown in Table \uppercase\expandafter{\romannumeral 12}. In this set of experiments, we used the optimal parameters explored for each task decoder. The results in the first row of Table  \uppercase\expandafter{\romannumeral 12}. represent the experimental results of our baseline Panoformer; the second row represents the performance when we used the proposed background depth resolving component in conjunction with BGDNet's background replacement strategy; and the third and fourth rows represent the performance when we combined our proposed fusion mask generation and adaptive fusion components, respectively. From the results in Table  \uppercase\expandafter{\romannumeral 12}., it is clear that the core of our framework's performance improvement lies in the three task components rather than multi-task learning, and the fusion mask generation component shows the most significant improvement. This experimental result also fully validates the core argument of this paper: to make the background depth constraint more reasonable, we need to more carefully identify which regions in the input indoor panorama image belong to the background.

In summary, our discussion validates both the proposed framework and the collaborative design of its subtasks.

\section{Limitation}
The proposed PAGCNet achieves significantly superior performance on current indoor depth estimation datasets. However, our framework still has several issues:

1) To adapt to more complex indoor scenes, we divide the scene into regular and irregular regions and only model the background for the regular regions. Therefore, we don't actually model the irregular regions, and naturally, we don't apply geometric constraints to their depth. This makes our framework perform poorly on some extremely irregular room structures, although these scenes are very rare.

2) In existing panoramic depth estimation datasets, the number of annotations for the room layout task cannot be matched one-to-one with the annotations for tasks like semantic segmentation and depth estimation. This mismatch forces us to train the room layout decoder separately before training other decoders. Therefore, our framework doesn't actually solve this problem of annotation imbalance between tasks.

\section{Conclusions}
This paper has proposed a pose-aware and geometry-constrained framework for irregular panoramic depth estimation.Our framework first employs multiple task-specific decoders to jointly estimate room layout, camera pose, depth, and region segmentation from a single input panorama. A pose-aware background depth resolving component uses tasks decoder's prediction to refine the camera pose and subsequently uses the camera pose to compute the background depth of regular enclosed regions, which serves as a strong geometric prior. Based on the output of the region segmentation decoder, a fusion mask generation component produces a fusion weight map to guide where and to what extent the geometry-constrained background depth should correct the depth decoder's prediction. Finally, an adaptive fusion component integrates this refined background depth with the initial depth prediction, guided by the fusion weight. Extensive experiments on Matterport3D, Structured3D, and Replica datasets demonstrate that our method achieves significantly superior performance compared to current open-source methods.

\bibliographystyle{IEEEtran}
\normalem
\bibliography{reference}{}

\vspace{-1.5cm}
\begin{IEEEbiography}[{\includegraphics[width=1in,height=1.25in,clip,keepaspectratio]{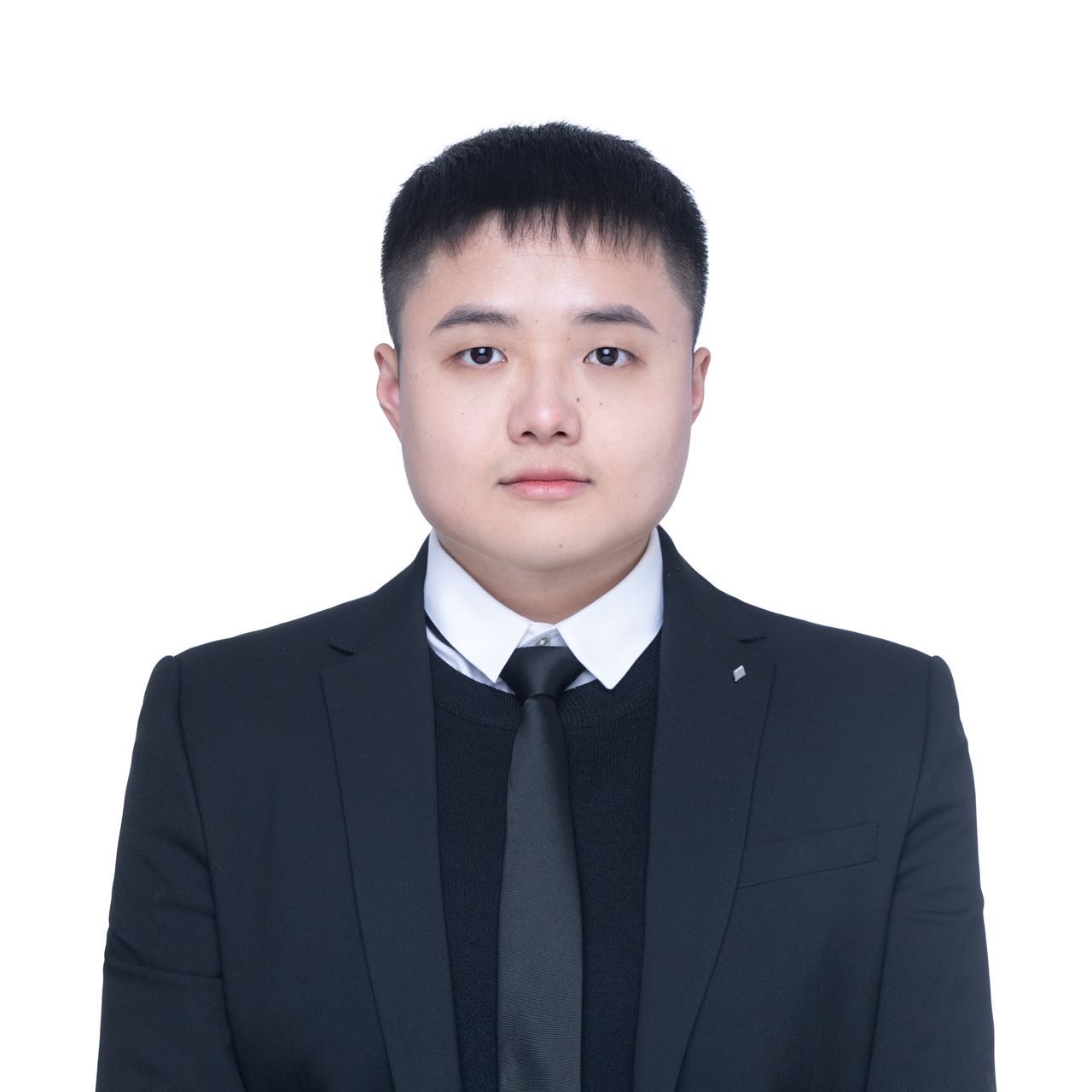}}]
{Kanglin Ning} received a B.S. degree from the Dalian University of Technology, Dalian, China, in 2016. and received the M.S. degree from the Department of Computer Science and Technology, the High-tech Institute of Xi'an. He is currently working toward a Ph.D. degree from the School of Computer Science, Harbin Institute of Technology (HIT), Harbin, China. His research interests include image processing, computer vision, depth estimation, object detection, and 3D object detection.
\end{IEEEbiography}

\vspace{-1.5cm}
\begin{IEEEbiography}[{\includegraphics[width=1in,height=1.25in,clip,keepaspectratio]{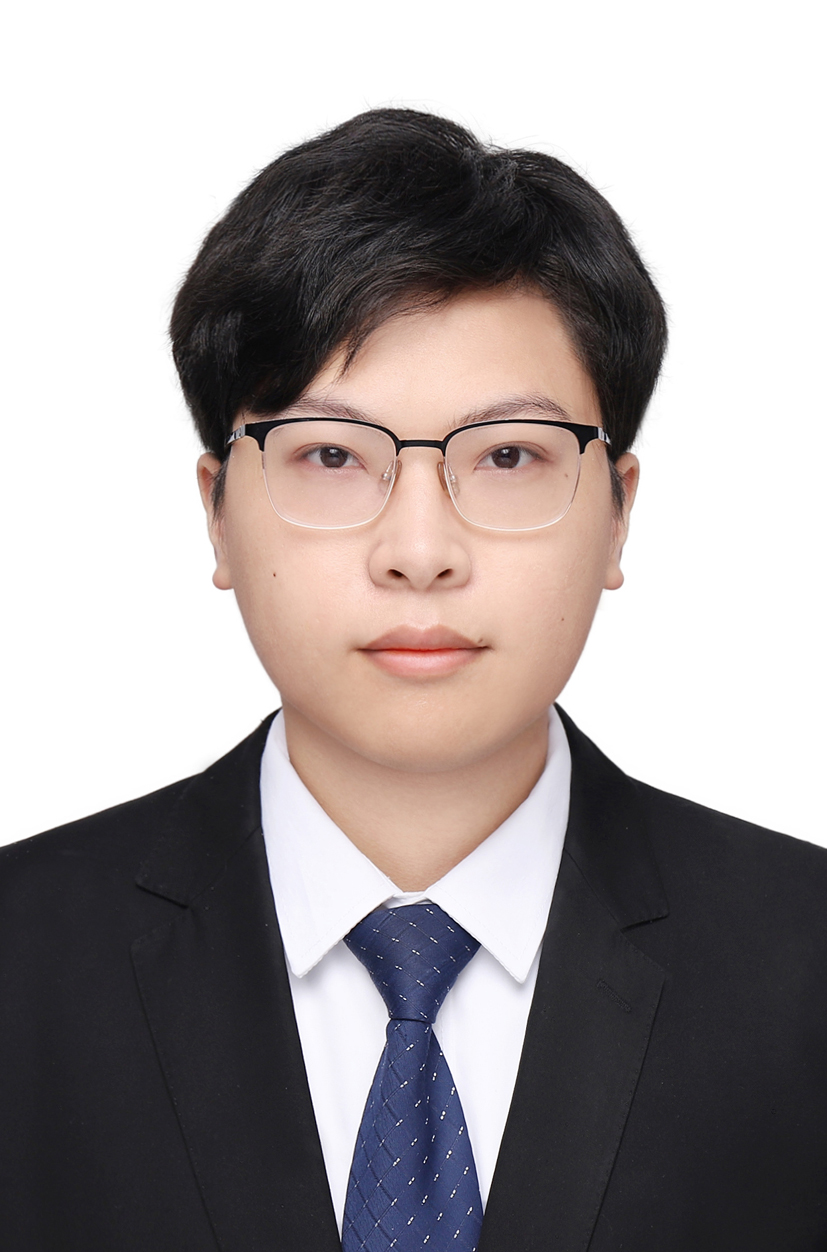}}]
{Ruzhao Chen} received a B.S. degree from the University of Electronic Science and Technology of China, Cheng Du, China, in 2023. He is currently working toward a M.S. degree from the School of Computer Science, Harbin Institute of Technology (HIT), Harbin, China. His research interests include image processing, computer vision, depth estimation.
\end{IEEEbiography}

\vspace{-1.5cm}
\begin{IEEEbiography}[{\includegraphics[width=1in,height=1.25in,clip,keepaspectratio]{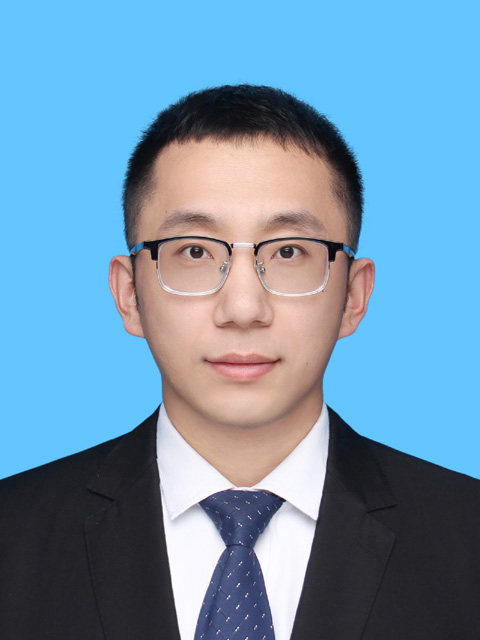}}]
{Penghong Wang} received the M.S. degrees from Computer Science and Technology, Taiyuan University of Science and Technology, Taiyuan,  China, in 2020, and received the Ph.D. degree in 
computer science from HIT, Harbin, China, in 2024. From 2021 to 2023, he was with Peng Cheng Laboratory. He is currently a postdoc with the School of Computer Science and Technology, HIT. His main research interests include wireless sensor networks, semantic communication and computer vision.
\end{IEEEbiography}

\vspace{-1.5cm}
\begin{IEEEbiography}[{\includegraphics[width=1in,height=1.25in,clip,keepaspectratio]{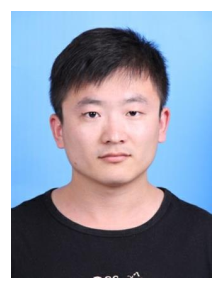}}]
{Xingtao Wang} received his B.S. degree from the Harbin Institute of Technology (HIT), Harbin, China, in 2016, and received the Ph.D. degree in computer science from HIT, Harbin, China, in 2022. From 2020 to 2022, he was with Peng Cheng Laboratory. He is currently a postdoc with the School of Computer Science and Technology, HIT. His research interests include point cloud denoising, mesh denoising, and deep learning.
\end{IEEEbiography}

\vspace{-1.5cm}
\begin{IEEEbiography}[{\includegraphics[width=1in,height=1.25in,clip,keepaspectratio]{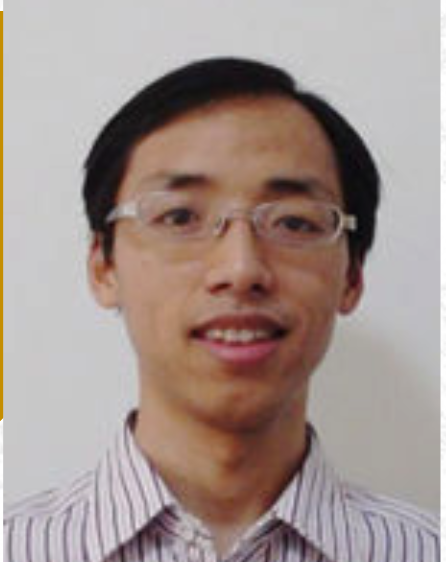}}]
{Ruiqin Xiong} g (M’08–SM’17) received the B.S. degree from the University of Science and Technology of China, Hefei, China, in 2001, and the Ph.D. degree from the Institute of Computing Technology, Chinese Academy of Sciences, Beijing, China, in 2007.
From 2002 to 2007, he was a Research Intern with Microsoft Research Asia. From 2007 to 2009, he was a Senior Research Associate with the University
of New South Wales, Australia. He joined the School of Electronic Engineering and Computer Science, Institute of Digital Media, Peking University, in 2010, where he is currently a Professor. He has authored over 110 technical papers in referred international journals and conferences. His research interests include statistical image
modeling, deep learning, and image and video processing, compression, and communications. He received the Best Student Paper Award from the SPIE
Conference on Visual Communications and Image Processing 2005, and the Best Paper Award from the IEEE Visual Communications and Image Processing 2011. He was also a co-recipient of the Best Student Paper Award at the IEEE Visual Communications and Image Processing 2017.
\end{IEEEbiography}

\vspace{-1.5cm}
\begin{IEEEbiography}[{\includegraphics[width=1in,height=1.25in,clip,keepaspectratio]{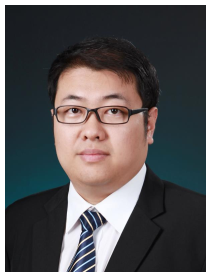}}]
{Xiaopeng Fan} (S'07-M'09-SM'17) received the B.S. and M.S. degrees from the Harbin Institute of Technology (HIT), Harbin, China, in 2001 and 2003, respectively, and the Ph.D. degree from The Hong Kong University of Science and Technology, Hong
Kong, in 2009. He joined HIT in 2009, where he is currently a Professor. From 2003 to 2005, he was with Intel Corporation, China, as a Software Engineer. From 2011 to 2012, he was with Microsoft Research Asia as a Visiting Researcher. From 2015 to 2016, he was with the Hong Kong University of Science and Technology as a Research Assistant Professor. He has authored one book and more than 100 articles in refereed journals and conference proceedings. His current research interests include video coding and transmission, image processing, and computer vision. He served as a Program Chair for PCM2017, Chair for IEEE SGC2015, and Co-Chair for MCSN2015. He was an Associate Editor of IEEE 1857 Standard in 2012. He received Outstanding Contributions to the Development of IEEE Standard 1857 by IEEE in 2013.
\end{IEEEbiography}

\end{document}